\documentclass[conference]{IEEEtran}
\usepackage{hyperref}
\hypersetup{
 unicode=true,
 pdfstartview={FitH},
 colorlinks=true,
 citecolor=blue,
 linkcolor=blue,
 urlcolor=blue
}

\usepackage[linesnumbered,algoruled]{algorithm2e}
\usepackage{algpseudocode}

\usepackage{caption}
\usepackage{subcaption}
\usepackage{multirow}
\usepackage{makecell}
\usepackage[inline]{enumitem}
\usepackage{amssymb}
\usepackage{threeparttable} 
\usepackage{listings}
\usepackage[comments]{annotations}

\newcommand{\nop}[1]{}

\newcommand{\Com}[1]{}

\usepackage{listings}
\usepackage{pifont}

\crefname{algocf}{alg.}{algs.}
\Crefname{algocf}{Algorithm}{Algorithms}
\SetKwRepeat{Do}{do}{while}

\usepackage{enumitem}
\usepackage{acronym}
\usepackage{booktabs}
\newcolumntype{L}[1]{>{\raggedright\let\newline\\\arraybackslash\hspace{0pt}}m{#1}}
\newcolumntype{C}[1]{>{\centering\let\newline\\\arraybackslash\hspace{0pt}}m{#1}}
\newcolumntype{R}[1]{>{\raggedleft\let\newline\\\arraybackslash\hspace{0pt}}m{#1}}

\usepackage{amsmath,amsfonts}
\usepackage{url}

\usepackage{graphicx, float}

\usepackage{xcolor}
\usepackage{tcolorbox}
\tcbuselibrary{skins, breakable}

\newtcolorbox{promptbox}[1][]{
    colback=gray!10, 
    colframe=gray!50, 
    arc=2pt, 
    boxrule=1pt, 
    fontupper=\footnotesize\ttfamily,  
    boxsep=2pt,                 
    left=3pt,                   
    right=3pt,                  
    top=2pt,                    
    bottom=2pt,                 
    title=\textbf{HLS Agent System Prompt}, 
    fonttitle=\sffamily\footnotesize,
    colbacktitle=black!70,
    enhanced,
    breakable,
    before upper={\parindent0pt}, 
    #1 
}

\newtcolorbox{excerptWithLatency}[2][]{
    colback=blue!14!black!7,
    colframe=blue!40!black!90,
    colbacktitle=blue!25!black!90,
    coltitle=white,
    arc=2pt,
    boxrule=1pt,
    fontupper=\footnotesize\ttfamily,
    boxsep=2pt,
    left=3pt,
    right=3pt,
    title=Excerpt from Reasoning Tokens: \textbf{#2},
    fonttitle=\sffamily\footnotesize,
    enhanced,
    before upper={\parindent0pt\vspace{-3pt}},
    unbreakable,
    overlay unbroken and first={
        \draw[black!30, line width=0.8pt] 
            (segmentation.west) -- (segmentation.east);
    },
    lower separated=false,
    before lower={\vspace{-4pt} },
    after lower={\vspace{-5pt}},
    after upper={\vspace{-2pt}},
    fontlower=\footnotesize\itshape,
    #1
}

\newtcolorbox{excerpt}[2][]{
    colback=blue!14!black!7,
    colframe=blue!40!black!90,
    colbacktitle=blue!25!black!90,
    coltitle=white,
    arc=2pt,
    boxrule=1pt,
    fontupper=\footnotesize\ttfamily,
    boxsep=2pt,
    left=3pt,
    right=3pt,
    title=Excerpt from Reasoning Tokens: \textbf{#2},
    fonttitle=\sffamily\footnotesize,
    enhanced,
    before upper={\parindent0pt\vspace{-3pt}},
    after upper={\vspace{-2pt}},
    unbreakable,
    #1
}

\begin{document}
\IEEEoverridecommandlockouts

\title{Can Reasoning Models Reason about Hardware? An Agentic HLS Perspective
}

\author{Luca Collini~\IEEEmembership{Graduate~Student~Member,~IEEE}, Andrew Hennessee, Ramesh Karri~\IEEEmembership{Fellow,~IEEE}, \\
Siddharth Garg~\IEEEmembership{Member,~IEEE}
\thanks{All authors are with the NYU Tandon School of Engineering, Brooklyn, NY 11201 (e-mail: \{lc4976, ajh9498@nyu.edu, rkarri, sg175\}@nyu.edu).}%
}


\maketitle

\begin{abstract}
Recent Large Language Models (LLMs) such as OpenAI o3-mini and DeepSeek-R1 use enhanced reasoning through Chain-of-Thought (CoT).
Their potential in hardware design, which relies on expert-driven iterative optimization, remains unexplored. This paper investigates whether reasoning LLMs can address challenges in High-Level Synthesis (HLS) design space exploration and optimization. 
During HLS, engineers manually define pragmas/directives to balance performance and resource constraints. We propose an LLM-based optimization agentic framework that automatically restructures code, inserts pragmas, and identifies optimal design points via feedback from HLS tools and access to integer-linear programming (ILP) solvers. Experiments compare reasoning models against conventional LLMs on benchmarks using success rate, efficiency, and design quality (area/latency) metrics, and provide the first-ever glimpse into the CoTs produced by a powerful open-source reasoning model like DeepSeek-R1. 
\end{abstract}


\section{Introduction}




Large Language Models (LLMs) are rapidly improving in their capabilities. Recently introduced models, notably OpenAI o3‑mini and DeepSeek-R1, leverage Chain-of-Thought (CoT) mechanisms to ``reason" about user prompts, achieving state-of-the-art results in challenging problem domains like mathematics, coding, and scientific benchmarks~\cite{deepseekr1}. CoT is a prompt engineering technique that encourages LLMs to break down problems into intermediate steps.
These models undergo specialized training that emphasizes the development of reasoning skills. For instance, DeepSeek's R1 model has been trained on extensive datasets that are rich in mathematical and logical content. Some models incorporate system-level prompts or architectural features that guide their behavior to further enhance autonomous reasoning. 

This brings us to the motivating question: \emph{Can reasoning models (\emph{e.g.}, DeepSeek-R1, o3-mini) reason about hardware design, and how do they compare to non-reasoning models (\emph{e.g.}., DeepSeek-V3)?} 
Despite access to powerful electronic design automation (EDA) tools, many design tasks 
rely on engineer expertise. For instance, system architects reason about the performance of a design consisting of multiple IP blocks/modules, and allocate resources to optimize performance
based on intuition, experience, or using optimizers like integer linear programming (ILP) solvers. 

In this paper, we use high-level synthesis (HLS) to explore the reasoning capabilities of state-of-art LLMs.
HLS starts with a C/C++ specification and uses pragmas and directives to express design optimizations. 
While translation from a C/C++ design with pragmas to hardware is automated, 
HLS does not tackle the task of identifying pragmas and directives for the best performance under a resource constraint. System architects do this task, exploring solutions until they are satisfied with the result. Design space exploration tools can yield good results but require orders of magnitude more samples than those required by an expert. 

This paper is the \emph{first} to explore
whether recent reasoning models can aid or even replace system architects. To this end, our novel contributions are:
\begin{itemize}[noitemsep, topsep=0pt]
\item we  \href{https://anonymous.4open.science/r/anonymous-sub-23F5/README.md}{open source} an \emph{automated}, \emph{agent-based} HLS optimization flow that automatically rewrites code, inserts pragmas and performs full-system optimization to minimize 
latency within a total area constraint. 
\item Our results on reasoning and baseline models demonstrate that while the proposed agentic workflow improves on the current state-of-the-art (sometimes even matching human performance), reasoning models are comparable with non-reasoning counterparts on hardware optimization. 
\item We provide the first glimpse into the chains-of-thought (CoT) produced by a reasoning model (DeepSeek-R1) in the hardware context, revealing both the promise and key gaps in its reasoning capabilities for hardware tasks. Hardware optimization remains a challenging benchmark for even the most powerful LLMs.
\end{itemize}
We begin by covering some necessary background. 
\section{Background}
\subsection{High-Level Synthesis (HLS)}
HLS tools transform high-level code (e.g., C/C++) into a register-transfer level (RTL) implementation. 
Pragmas and directives allow designers to guide and explore microarchitectural trade-offs (e.g., parallelism, resource sharing) directly within the high-level code, balancing performance, area, and power constraints. The design space of possible pragma optimization grows exponentially as the design complexity increases. Designers use their knowledge to navigate this complex design space and find solutions that satisfy their constraints. Researchers have proposed numerous solutions to perform design space exploration~\cite{GNN_OPT, HLS_opt, HLS_OPT2} using both classic and ML-based DSE approaches. These solutions require a high number of syntheses and are often domain-specific. 

Another approach is to identify possible solutions for the subkernels (\emph{e.g.,} by spanning the unrolling factors) and use optimizers like integer linear programming (ILP) solvers to identify the best subkernel composition~\cite{ilp1,ilp2,ilp3}. 
In this work, we explore the capabilities of reasoning models to approach this problem, evaluating the problem formulation and the process (number and quality of formulated ILP problems, number of syntheses, PPA, cost, and time) to explore whether generative models follow a process similar to an engineer.

\subsection{Large Language Models (LLMs)} 

LLMs are trained on massive amounts of text data and excel at tasks such as code generation and translation, particularly in languages such as C, C++, and Python~\cite{NEURIPS2023_43e9d647}. However, their performance declines when working with Hardware Description Languages (HDLs), such as Verilog or VHDL, due to the limited amount of training data available in those languages~\cite{hammond2022}.
Recent LLMs, such as OpenAI o3-mini and DeepSeek-R1, have reasoning capabilities~\cite{deepseekr1}. Unlike traditional LLMs, which are trained using supervised learning on massive datasets, reasoning-enhanced LLMs integrate Chain-of-Thought prompting, supervised fine-tuning on reasoning tasks, and reinforcement learning to improve logical reasoning. During inference, rather than providing a final answer, these LLMs generate intermediate reasoning steps, mimicking human-like problem-solving. This approach improves performance, particularly in reasoning tasks such as coding, mathematics, science, and logical reasoning~\cite{google-CoT-2022}.

LLM-based frameworks have been proposed to implement Verilog designs~\cite{autochip, evaluating_llms_4_verilog,rome}. The design complexity is limited by the LLMs' capabilities in Verilog generation. For this reason, LLM frameworks for HLS have been proposed~\cite{c2hlsc,C2HLSC2,automatedhls,eval_llm_hls}, as LLMs perform better with high-level languages. 
The potential of new-generation reasoning models has not yet been explored in this domain. 
We propose an agent-based HLS optimization framework and explore whether LLMs can reason about design constraints, such as latency and area while refining solutions to meet performance targets.

\section{HLS Optimization Agent}
\begin{figure}[t]
    \centering
    \includegraphics[width=.96\columnwidth]{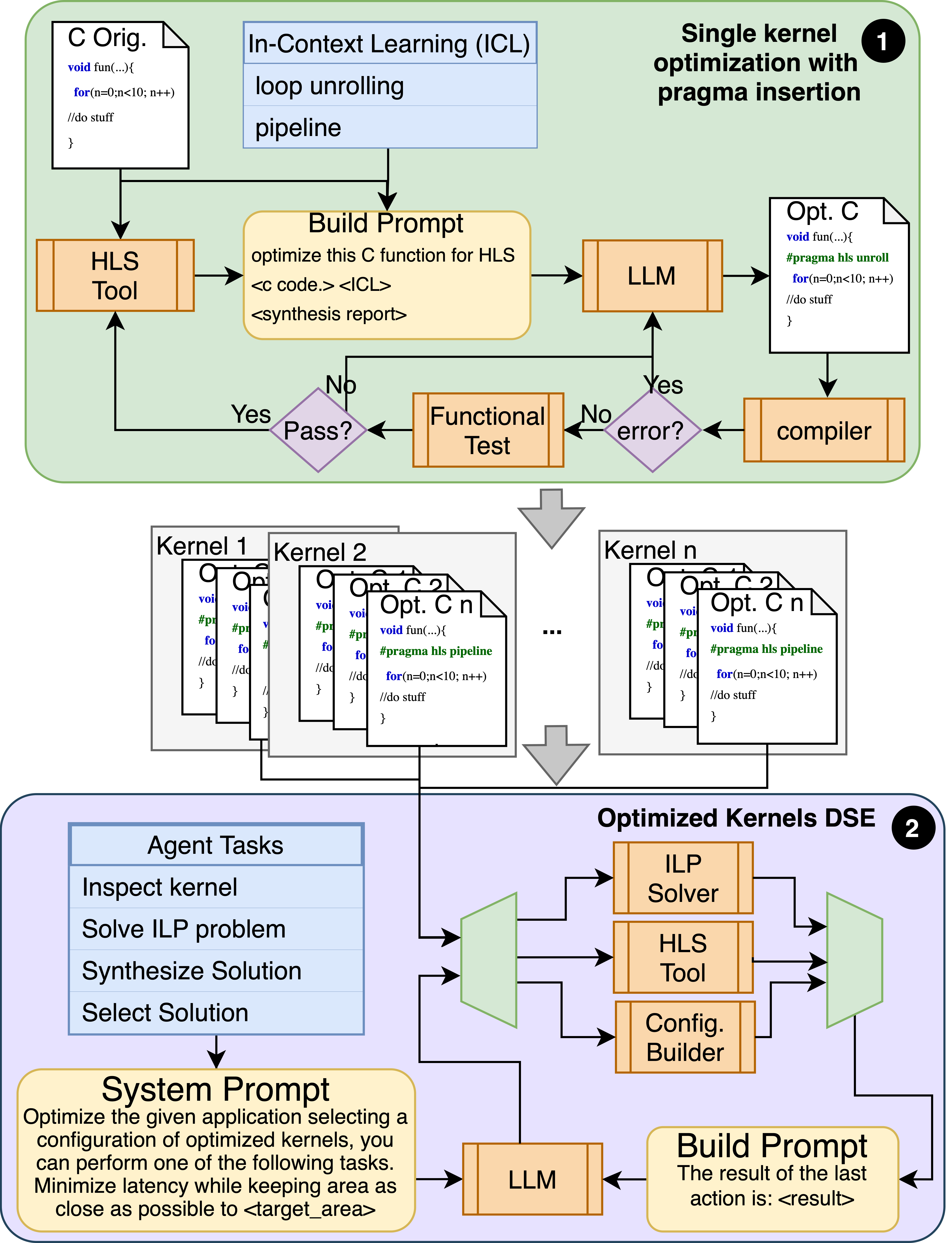}
    \caption{HLS agentic flow with two optimization tasks.}
    \label{fig:flow}
\vspace{-20pt}
\end{figure}

We frame two tasks to evaluate LLM reasoning:
    \ding{202}  \textit{Kernel-level pragma insertion and code transformations;} 
     \ding{203}  \textit{Full system optimization.}  
     We propose an agentic flow (as shown in \autoref{fig:flow}) to perform these tasks, giving the LLM access to tools (code inspection, system synthesis, and an ILP solver) to perform these tasks.

\paragraph{Kernel-Level Pragma Insertion and Code Transformations}  
The goal in Task~\ding{202} is to identify different implementations spanning a range of area-performance tradeoffs for each kernel (C/C++ function) in the design via code transformations and pragma insertion. This is accomplished with an LLM in a feedback loop---in each iteration, the LLM is given the area, latency, and throughput result of the base implementation and the solutions identified so far and prompted to provide a new solution. 
We compile and run a functional test to ensure that the LLM does not inadvertently change the function (which was proven effective in~\cite{autochip,c2hlsc}) and use error feedback to prompt the LLM for bug fixes. Our agentic approach operates hierarchically, starting with the leaf-level functions and moving up to the 
top module. 
As shown in~\cite{c2hlsc}, working on multiple functions at a time often leads to hallucinations and errors. 
This bottom-up approach allows the LLM to focus on the function at hand. 
When traversing up the kernel hierarchy, we need to choose an implementation for each child function (from a previous iteration). We use a greedy approach to select the lowest latency option for each child. This approach does not guarantee an optimal solution, and the final solution might not meet the area target. For this reason, we keep every solution from task~\ding{202} for each function, which will be the input for task~\ding{203}.

\paragraph{Full System Optimization} 
Task~\ding{203} has the goal of optimizing the overall application, picking a configuration for each kernel in the design to minimize latency while keeping the area within a target constraint. These tasks are typically solved by system architects via trial-and-error, 
heuristic approaches or, as has been shown in prior work, using a formal optimizer like an ILP solver~\cite{ilp1,ilp2,ilp3}. 
We implement an LLM agent that can take one of four actions. 
\begin{enumerate*}
    \item Inspect Kernel, i.e., view its code or that of submodules (needed to minimize context length and token counts);
    \item Solve an Integer Linear Programming (ILP) Problem;
    \item Synthesize Solution, i.e., call the HLS tool with specified pragma values;
    \item Select Solution, i.e., use the solution returned by the ILP solver.
\end{enumerate*}
For the ILP problem formulation, we prompt the LLM to provide a Python script using the Google OR-Tools library. At each iteration, the LLM provides an action. We parse the response, call the respective tool, and report the outcome to the LLM agent. The task ends when the agent selects a solution. 
Crucially, system optimizations require an understanding of hardware; for example,  
the latency of parallel modules is the \emph{max} of the two latencies and not their \emph{sum} 
as would be the case in sequential software execution. This plays an important role in being 
able to formulate accurate ILP formulations.

\section{Experimental Evaluation}

\subsection{Experimental Setup}
\textbf{}We implemented the agentic flow  in \autoref{fig:flow} using Python. We used Catapult HLS and target nandgate45 library. The two tasks are implemented as two separate conversations with the LLM. This reduces token use and avoids exceeding the context size in the second task. 
The overall flow is implemented in fifteen hundred lines of Python and is fully automated.
Code, benchmarks, and raw results are provided \href{https://anonymous.4open.science/r/anonymous-sub-23F5/README.md}{here}.

Below is a shortened version of our system prompt; our system and user prompt templates are available in our source code. To avoid biasing the models, we did not provide additional instructions or examples of how a designer might approach the task (e.g., inspect function dependencies, formulate an ILP model, evaluate results).
\begin{promptbox}
\texttt{You are an HLS Optimization Agent tasked with optimizing a C application accelerated using High-Level Synthesis.\\
Your goal is to find the best combination of function options that minimize latency while keeping the total {constraint} as close as possible to a target value. \\
At every iteration, you have four options:
<options> \\
Only reply with one of the five options following the format provided.
}
\end{promptbox}

Our choices of LLMs are as follows: \textbf{DeepSeek-V3} serves as our baseline LLM without reasoning. \textbf{DeepSeek-R1} and \textbf{OpenAI o3-mini} are the reasoning models. o3-mini is the latest reasoning model released by OpenAI, offering a faster response time and a lower cost compared to the previous o1 reasoning model. DeepSeek-R1 is a newly released reasoning model that provides CoT reasoning tokens, which o3-mini does not. This feature allows for a deep dive into the reasoning behind the implemented agent actions (\autoref{sec:deep}).

\autoref{tab:benchmarks} presents the benchmarks and their characteristics. We created six synthetic benchmarks that implement different control and data dependency graphs including compositions of sequential and parallel modules
to evaluate whether LLMs can reason about latency and area from a hardware perspective. 
These benchmarks also mitigate concerns about data contamination as they could not have 
been part of any training data. Additionally, we selected six real-world benchmarks. 

\setlength{\tabcolsep}{2pt} 
\begin{table}[htb]
\caption{Benchmarks description and characterization.}
\label{tab:benchmarks}
\resizebox{\columnwidth}{!}{
\begin{tabular}{@{}llrrrrrrrr@{}}
\toprule
 &  & \#Kernels/ & \multicolumn{3}{c}{\# Lines} & \multicolumn{3}{c}{\# Operations} \\ \cmidrule{4-9} 
 
Benchmark & Description & \#Calls & Total & Min. & Max. & Total & Min. & Max. \\ \midrule
SYN1      & 2 sequential functions & 3/3 & 30 & 8 & 11 & 3 & 0 & 2 \\
SYN2      & 2 parallel functions & 3/3 & 31 & 9 & 11 & 4 & 1  & 2 \\
SYN3      & \begin{tabular}[c]{@{}l@{}}2 parallel pairs of \\ sequential functions\end{tabular} & 4/5 & 50 & 11 & 15 & 6 & 1 & 3 \\
SYN4      & \begin{tabular}[c]{@{}l@{}}2 parallel functions \\ followed by 2 \\ sequential functions\end{tabular} & 4/5 & 50 & 11 & 15 & 8 & 1 & 3 \\
SYN5      & SYN3 in a for loop & 4/5 & 54 & 11 & 17 & 8 & 1 & 3 \\
SYN6      & SYN4 in a for loop & 4/5 & 54 & 11 & 17 & 10 & 1 & 4 \\
AES       & AES Cipher core & 6/11 & 101 & 5 & 26 & 77 & 2 & 29 \\
Present   & Present Cipher core & 6/10 & 96 & 10 & 24 & 74 & 4 & 26 \\
SHA256    & SHA256 core & 2/2 & 70 & 17 & 53 & 127 & 11 & 116 \\
KMP       & \begin{tabular}[c]{@{}l@{}}KMP algorithm for\\ pattern searching\end{tabular} & 3/3 & 62 & 8 & 28 & 17 & 0 & 10 \\
FIR+IIR   & \begin{tabular}[c]{@{}l@{}}Runs FIR and IIR \\ in parallel on given\\ input stream\end{tabular} & 5/9 & 48 & 7 & 11 & 12 & 1 & 4 \\
NW        & \begin{tabular}[c]{@{}l@{}}Needleman-Wunsch \\ algorithm for sequence\\ alignment\end{tabular} & 4/5 & 109 & 6 & 52 & 73 & 5 & 37 \\ \bottomrule
\end{tabular}}
\end{table}
\setlength{\tabcolsep}{6pt} 
We ran the flow ten times for each benchmark, setting the number of optimized kernel options for task~\ding{202} to five. For each benchmark, we select an area target 10\% below the initial solution, allowing the LLM Agent to find a lower-area solution and then determine the minimum latency at that area.

\subsection{Implementation Challenges}\label{sec:challenges}

We encountered several challenges while implementing our proposed framework. Reasoning and non-reasoning models respond very differently to our prompts.
Our system prompt explicitly requests that explanations not be provided. 
Only the two reasoning models adhered to this instruction, replying with only the chosen action. In contrast, the chat model (DeepSeek-V3) included some explanation either before or after the chosen action. This made it more challenging to parse the agent's choice but provided insights into the chat model's decision-making process. 
We believe that because the reasoning models approach the problem in the Chain-of-Thought and provide the final answer only at the end, they comply with the required output structure while still following their statistical reasoning process. On the other hand, chat models do not have separate reasoning and output tokens. As a result, the explanations are, in some sense, their problem-solving process leading to the final answer. 

The official DeepSeek API experienced significant downtime during our experiments, so we 
additionally used a third-party provider for some of the DeepSeek-R1 experiments. Since we did not observe differences in the quality of results, we agglomerated the runs from first- and third-party APIs.

\subsection{Experimental Results}
We now present our experimental results, highlighting key observations from our empirical 
evaluations.
\begin{figure}[tb]
    \centering
    \begin{minipage}[b]{\columnwidth}
        \centering
        \hspace{0pt}\includegraphics[width=1\columnwidth]{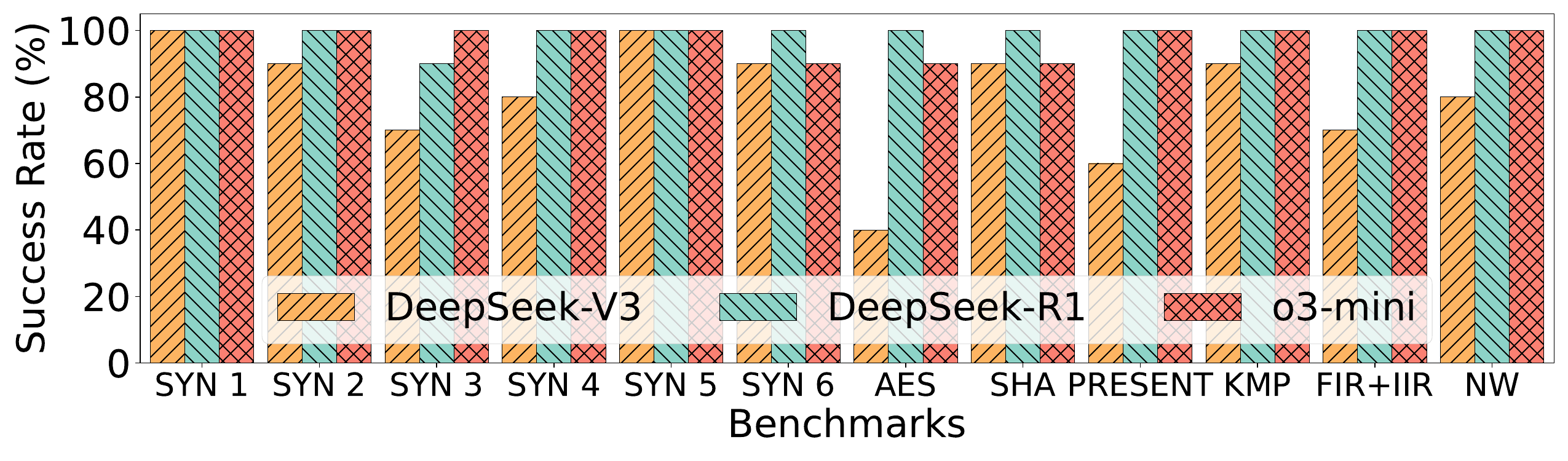}
        \vspace{-8pt}
        \label{fig:subfig1}
    \end{minipage}
    \begin{minipage}[b]{\columnwidth}
        \centering
        \hspace{0pt}\includegraphics[width=1\columnwidth]{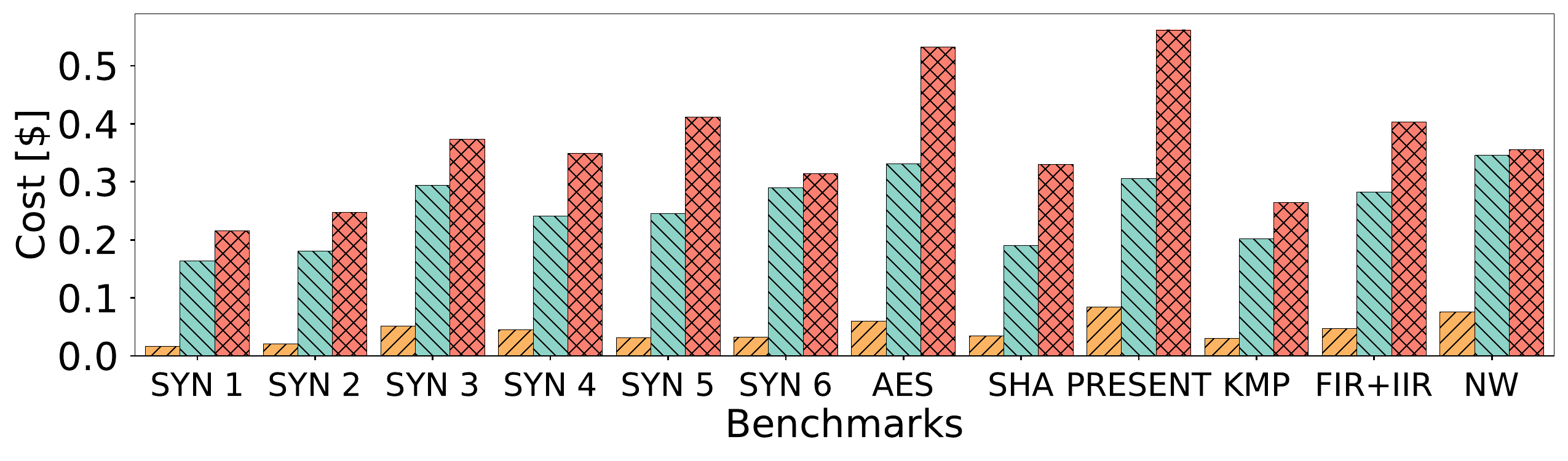}
        \vspace{-8pt}
        \label{fig:subfig2}
    \end{minipage}
    \begin{minipage}[b]{\columnwidth}
        \centering
        \hspace{0pt}\includegraphics[width=1\columnwidth]{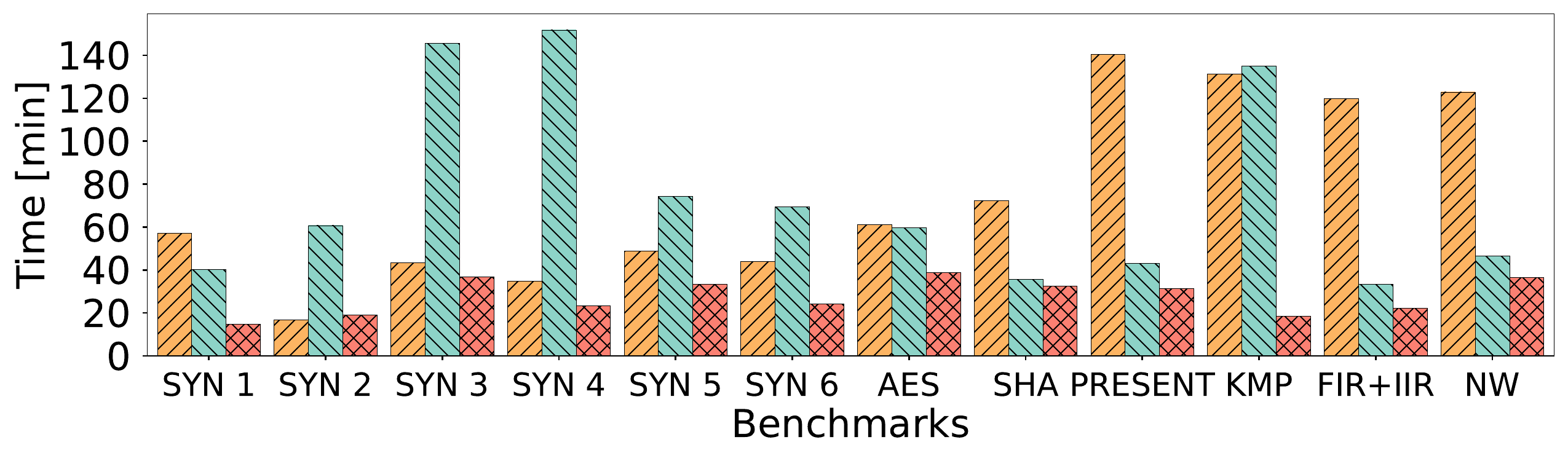}
        \vspace{-14pt}
        \label{fig:subfig3}
    \end{minipage}
    \caption{Success rate, cost, and runtime to run comparison between DeepSeek-V3, DeepSeek-R1, and o3-mini.}
    \label{fig:success}
    \vspace{-15pt}
\end{figure}

\paragraph{\textbf{Reasoning Models have Higher Success Rates But at Higher Cost}} \autoref{fig:success} presents the success rate, average cost, and average runtime for the twelve benchmarks across the ten runs.
Here, success means that the flow reaches the end, providing a result, regardless of whether the result meets the target area or not.

DeepSeek-V3 has the lowest success rate, performing its worst on AES with a success rate of 40\%. However, it still achieved a 100\% success rate on multiple benchmarks while maintaining the lowest cost (by at least a 4$\times$ margin) and a runtime often comparable to o3-mini. DeepSeek-V3 exhibits two distinct failure modes. The first, during task~\ding{202}, was due to modifying the kernel's functionality and being unable to repair it. The second, during task~\ding{203}, resulted from reaching the model's context size limit. o3-mini is the most expensive and fastest model. Its success rate is fairly consistent, and failures occur only sporadically due to unintended functionality modifications that it could not fix. DeepSeek-R1 is the slowest model, but it achieves the highest success rate, failing only once due to insertion of bugs during task~\ding{202}.  
The cost was calculated by multiplying the number of input and output tokens by the respective API costs. We used the full price provided without considering cached tokens or off-peak discounts (which would favor the lower-cost models). At the time of writing, the cost per million input/output tokens is \$0.27/\$1.10 for DeepSeek-V3, \$0.55/\$2.19 for DeepSeek-R1, and \$0.55/\$4.40 for o3-mini. Additionally, the DeepSeek models are open source, allowing for trade-offs between cost and performance depending on the chosen deployment options.

\begin{figure}[tb]
    \centering
    \begin{minipage}[b]{\columnwidth}
        \centering
        \hspace{0pt}\includegraphics[width=1\columnwidth]{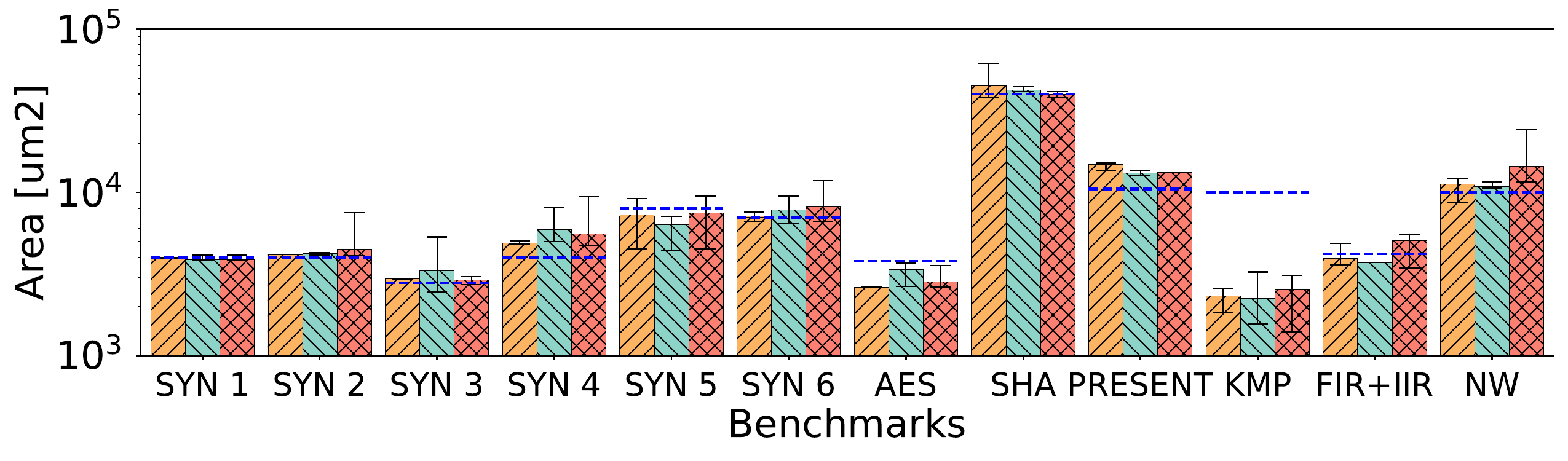}
        \vspace{-12pt}
        \label{fig:subfig4}
    \end{minipage}
    \begin{minipage}[b]{\columnwidth}
        \centering
        \hspace{0pt}\includegraphics[width=1\columnwidth]{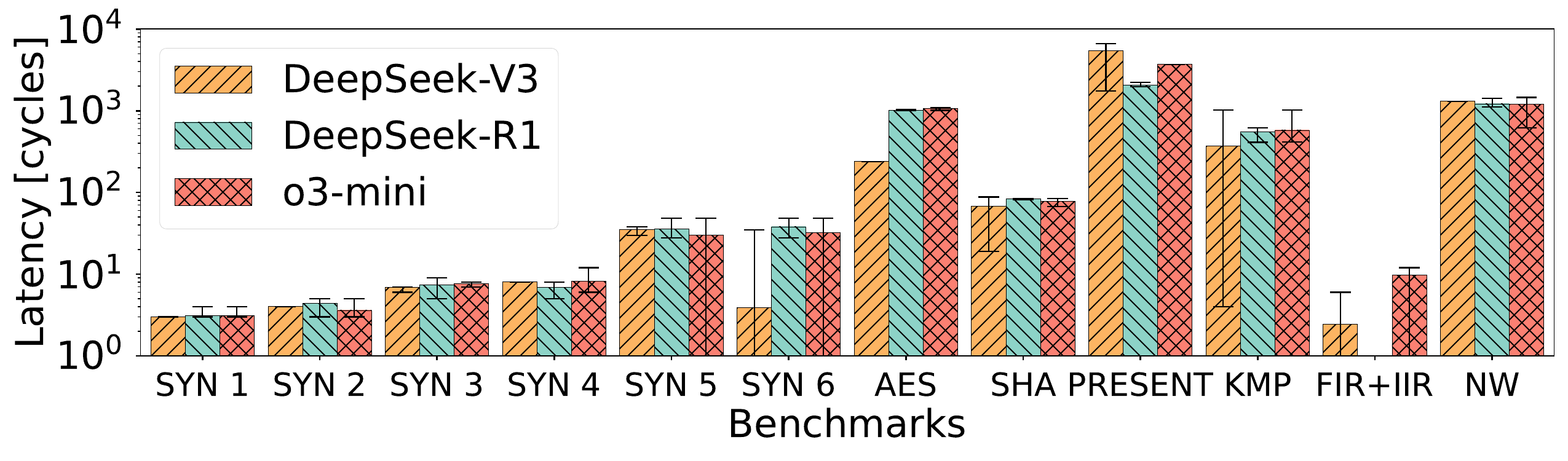}
        \vspace{-14pt}
        \label{fig:subfig5}
    \end{minipage}
    \caption{Synthesis result comparison between DeepSeek-V3, DeepSeek-R1, and o3-mini. The vertical ranges represent the min/max ranges. Log scales. The blue dash lines indicate the target area used for the benchmark.}
    \label{fig:synth_comp}
    \vspace{-10pt}
\end{figure}

\paragraph{\textbf{Synthesized Area-Latency Results are Comparable Across Models}}
\autoref{fig:synth_comp} presents synthesis results for all models and benchmarks. The bars represent the average values,  with minimum and maximum ranges annotated, and blue dashed lines show the target area for each benchmark. No model consistently outperforms all others---in fact, the non-reasonig DeepSeek-V3 model surpasses the reasoning ones in a few benchmarks. 
In three benchmarks (SYN2, SYN4, and Present) no model achieved the target area, although 
a human engineer can be reasonably expected to do so by simply picking the lowest area implementation of each module. 

\setlength{\tabcolsep}{4pt}
\begin{table}[htb]
\centering
\caption{Performance summary. $A$ and $L$ are the obtained area and latency.  ${tgt}$ and ${min}$ refer to target and minimum. }
\label{tab:scores}
\begin{tabular}{@{}lrrr|rrr|rrr@{}}
\toprule
 & \multicolumn{3}{c|}{$A\leq A_{tgt}$} & \multicolumn{3}{c|}{\begin{tabular}[c]{@{}c@{}} $A\leq A_{tgt}$ \&\&\\$L == L_{min}$\end{tabular}} & \multicolumn{3}{c}{\begin{tabular}[c]{@{}c@{}}$A>A_{tgt}$ \&\&\\ $A==A_{min}$\end{tabular}} \\ \cmidrule(l){2-10} 
Benchmark & \multicolumn{1}{c}{V3} & \multicolumn{1}{c}{R1} & \multicolumn{1}{c|}{o3-m} & \multicolumn{1}{c}{V3} & \multicolumn{1}{c}{R1} & \multicolumn{1}{c|}{o3-m} & \multicolumn{1}{c}{V3} & \multicolumn{1}{c}{R1} & \multicolumn{1}{c}{o3-m} \\ \midrule
SYN1         & 8     & \textbf{9}     & \textbf{9}     &8      &  \textbf{9}    & \textbf{9}     & - & - & -    \\
SYN2           &0      &  0    & 0   & - & - & -    &0      & 0     & \textbf{5}    \\
SYN3         & 0     & \textbf{3}     & \textbf{3}     &0      &  0    & \textbf{3}     & - & - & -    \\
SYN4           &0      &  0    & 0   & - & - & -    &0      & 0     & \textbf{1}    \\
SYN5         & 7     & \textbf{10}    & 7     &0      &  \textbf{4}    & 3     & - & - & -    \\
SYN6         & \textbf{5}     & \textbf{5}     & 1     &\textbf{5}      &  0    & 0     & - & - & -    \\
AES   & 4     & \textbf{10}    & 9     &\textbf{4}      &  0    & 0     & - & - & -    \\
SHA256    & 4     & 0     & \textbf{5}     &\textbf{3}      &  0    & 1     & - & - & -    \\
Present       &0      &  0    & 0  & - & - & -   &0      & \textbf{3}     & 0    \\
KMP      & 9     & \textbf{10}    & \textbf{10}    &\textbf{1}      &  0    & 0     & - & - & -    \\
FIR+IIR   & 5     & \textbf{10}    & 2     &5      &  \textbf{10}   & 2     & - & - & -    \\
NW  & \textbf{2}     & 0     & 0     &\textbf{2}      &  0    & 0     & - & - & -    \\\midrule
\textbf{Total} & 44 & \textbf{57} & 46 & \textbf{28} & 23 & 18 & 0 & 3 & \textbf{6} \\ \bottomrule
\vspace{-15pt}
\end{tabular}
\end{table}

\autoref{tab:scores} presents the number of times each model meets the target area for each benchmark. It also reports the score achieved for each benchmark, which is determined based on two scenarios: \begin{enumerate*}
    \item 
    A point is awarded each time a model meets the area target and achieves the lowest latency among the runs that meet the area target;
    \item When no model meets the target area, a point is awarded each time a model achieves the lowest area across the results.
\end{enumerate*}
From \autoref{tab:scores}, no model outperforms all others. DeepSeek-R1 has the highest number of runs that meet the area target, DeepSeek-V3 earns the most points in scenario 1, and o3-mini scores the most points in scenario 2.

\begin{figure}[htb]
    \vspace{-4pt}
    \centering
    \includegraphics[width=0.9\columnwidth]{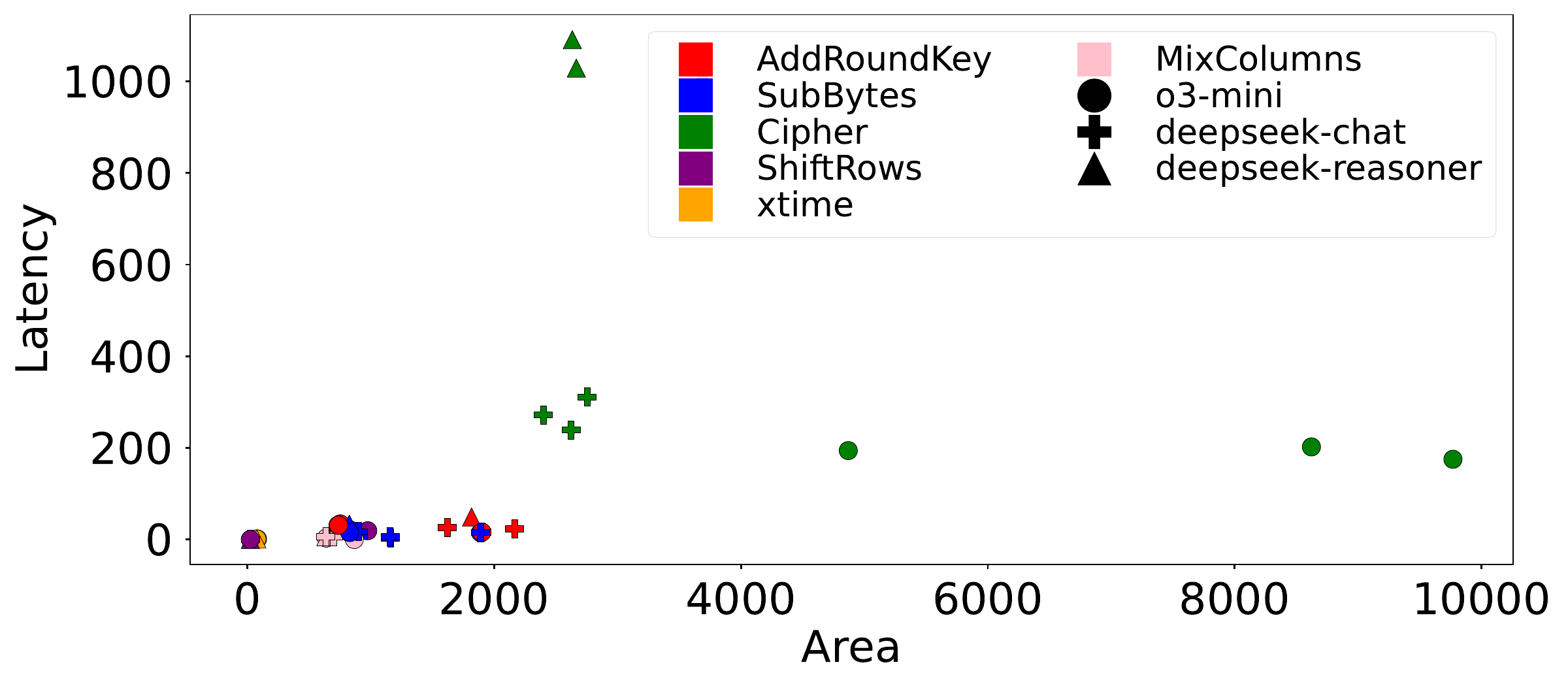}
    \vspace{-8pt}
    \caption{Solutions for AES sub-kernels for each model.}
    \label{fig:aes}
    \vspace{-8pt}
\end{figure}

\paragraph{\textbf{Performance Breakdowns by Task}}
Our evaluations thus far do not distinguish whether the achieved result quality is the result of a well-performed task~\ding{202} or task~\ding{203}. If the model selects a strong (weak) set of optimized kernels, task~\ding{203} becomes easier (harder). 
\autoref{fig:aes} shows the latency and area for each AES subkernel optimized in task~\ding{202} by each model. Interestingly, we find that the 
non-reasoning model produced better designs for Cipher, the largest submodule in AES.

To better evaluate the models on task~\ding{203}, we reran it in isolation for the AES benchmark, starting from the same subkernel implementations for all models. In this setting, \emph{DeepSeek-R1 performed the best}, achieving an area of 3596 $\mu m^2$ (the area target was 3800$\mu m^2$) and a latency of 736 cycles. DeepSeek-V3's best result reached an area of 3707$\mu m^2$ and a latency of 763 cycles. o3-mini's best result had an area of 4339$\mu m^2$ (failing to meet the target area) and a latency of 638 cycles. These results significantly differ from the best overall results, underscoring the impact of task~\ding{202} on achieving strong performance in task~\ding {203}.

\paragraph{\textbf{All Models Struggle to Formulate ILPs Correctly, but Show Promise}}
Focusing on task~\ding{203}, we evaluate the models' ability to formulate integer linear programming (ILP) problems for optimizing subkernel selection. 
An ILP problem formulation for this task consists of three parts:
\begin{itemize}
    \item Defining 1-hot binary variables for each subkernel option;
    \item Formulating the area and latency models;
    \item Formulating the optimization objective and constraints.
\end{itemize}
All models \emph{correctly} define 1-hot binary variables,  but differ in how they treat objectives and constraints. All models struggle to model latency.  

DeepSeek-V3 constrains the area while minimizing latency. Interestingly,  
DeepSeek-R1 and o3-mini opt for a multi-objective formulation: $\min(\alpha\cdot latency + |area-target|)$, using a
Lagrange multiplier for area instead of explicitly constraining it.
Setting the area target as a constraint causes the solver to fail if the target is unreachable. When this occurs, the model increases the limit and retries until a solution is found. This approach is inefficient and often leads to DeepSeek-V3 exhausting its context tokens and failing.

The latency model formulation task proved challenging. Kernel latency depends on the structure of the data flow graph, and parallelism can be inferred by looking at dependencies in the code. \autoref{tab:latency_models} presents data flow graphs, correct latency models, and the latency models generated by the LLMs.

\setlength{\tabcolsep}{2pt}
\def\arraystretch{1.1}
\begin{table}[htb]
\centering
\resizebox{\columnwidth}{!}{
\begin{threeparttable}

\caption{Latency formulations by models for benchmarks. For AES-- \underline{A}dd \underline{K}ey; \underline{S}hift \underline{R}ow; \underline{S}ubstitution \underline{B}ox; \underline{M}ix \underline{C}olumn. For NeedWun-- \underline{F}ill \underline{M}atrix; \underline{T}race\underline{B}ack; \underline{R}everse \underline{S}tring.}
\label{tab:latency_models}
\vspace{-2pt}
\begin{tabular}{@{}c|l|l@{}}
\toprule
Data Flow Graph & \multicolumn{1}{c|}{Latency Formulations} & \multicolumn{1}{c}{Models} \\ 
\midrule
\multirow{5}{*}[3pt]{\includegraphics[width=0.38\columnwidth]{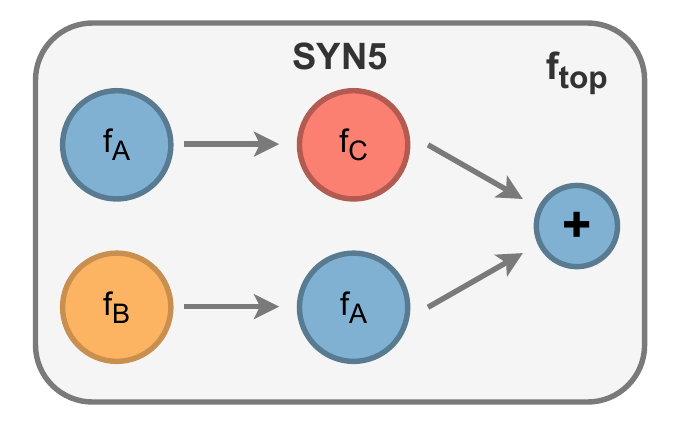}} 
&\ding{192} \quad $f_{\text{top}}$ & R1 \\
&\ding{193} \quad $f_{\text{top}} + f_A + f_B + f_C$ & R1, o3-m \\ 
&\ding{194} \quad $f_{\text{top}} + 2f_A + f_B + f_C$ & R1 \\ 
&\ding{195} \quad $\max(f_{\text{top}}, f_A, f_B, f_C)$ & o3-m \\ 
\cmidrule{2-3}
& \multicolumn{2}{l}{\quad\textbf{Correct:} $f_{\text{top}} + \max(f_A+f_C,f_B+f_A)$} \\ 
\midrule
\multirow{5}{*}[3pt]{\includegraphics[width=0.38\columnwidth]{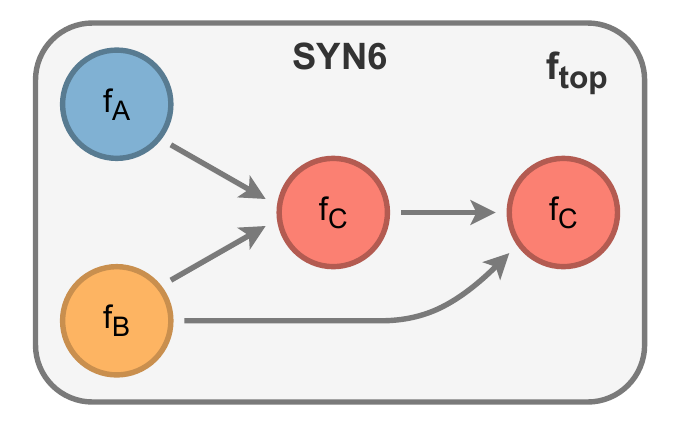}} 
&\ding{192} \quad $f_{\text{top}}$ & R1, o3-m \\
&\ding{193} \quad $f_{\text{top}} + f_A + f_B + f_C$ & V3, R1 \\
&\ding{194} \quad $f_{\text{top}} + f_A + f_B + 2f_C$ & R1 \\
&\ding{195} \quad $f_{\text{top}} + \max(f_A, f_B, f_C)$ & R1, o3-m \\ 
\cmidrule{2-3}
& \multicolumn{2}{l}{\quad\textbf{Correct:} $f_{\text{top}} + \max(f_A, f_B) + 2f_C$} \\ 
\midrule
\multirow{5}{*}[3pt]{\includegraphics[width=0.38\columnwidth]{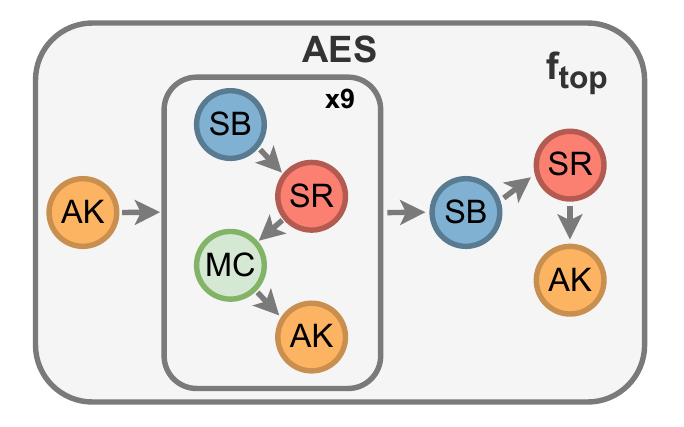}} 
&\ding{192}  $f_{\text{top}}+AK+SB+SR+MC$ & R1, o3-m \\
&\ding{193}  $f_{\text{top}}+3AK+SB+SR+MC $ & R1, o3-m \\ 
&  &  \\
&  &  \\
\cmidrule{2-3}
& \multicolumn{2}{l}{\quad\textbf{Correct:} $f_{\text{top}}+AK+SB+SR+MC$ \tnote{*}} \\ 
\midrule
\multirow{5}{*}[3pt]{\includegraphics[width=0.38\columnwidth]{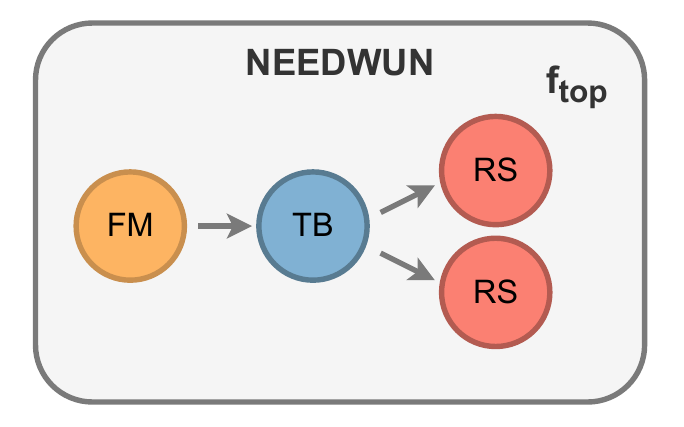}} 
&\ding{192}\quad  $f_{\text{top}}$ & o3-m \\
&\ding{193}\quad  $FM+TB$ & R1, o3-m \\ 
&\ding{194}\quad  $f_{\text{top}}+FM+TB+RS$ & V3, R1 \\ 
&  &  \\
\cmidrule{2-3}
& \multicolumn{2}{l}{\quad\textbf{Correct:} $f_{\text{top}}+FM+TB+RS$} \\ 
\bottomrule
\end{tabular}

\begin{tablenotes}
    \item[*] This is a good approximation.
\end{tablenotes}
\end{threeparttable}
}\vspace{-8pt}
\end{table}
\setlength{\tabcolsep}{6pt}
The results show that the models struggle to infer parallelism in the kernels. The only correct ILP formulations occur in cases where no parallelism exists between subkernels, such as in AES, or by chance, as in Needleman-Wunsch. By default, the models either consider only $f_{\text{top}}$ (the top-level function latency), or sum all 
latencies, sometimes factoring in the number of function calls.

\begin{figure}[tbh]\hspace{0pt}
\vspace{-12pt}
\includegraphics[width=1\columnwidth]{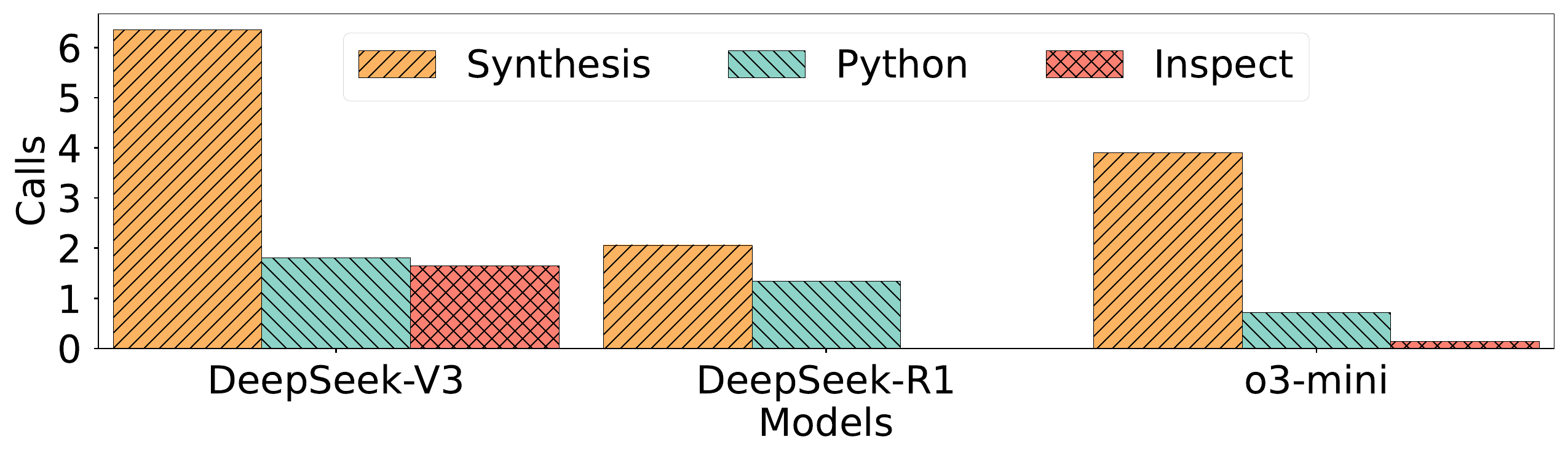}
    \vspace{-10pt}
    \caption{Average \# actions across all benchmarks for each LLM.}
    \label{fig:actions}
    \vspace{-0.3in}
\end{figure}

\paragraph{\textbf{Non-reasoning Model Call more Actions}}
\autoref{fig:actions} presents the average number of actions taken by each model across all benchmarks.

From the plots, DeepSeek-V3 has the highest average number of actions, while
reasoning models follow a direct approach, potentially because 
they rely more on ``learned" behavior. As an example, 
DeepSeek-R1 never opted to inspect code (note that it does have access to various latency-area points for each submodule). 
However, not inspecting code limits its ability to accurately model latency in ILP problems since it cannot identify dependencies between modules. 

\subsection{A Deep Dive into Reasoning}\label{sec:deep}
In this section, we examine LLM reasoning, including decision-making processes and the motivations behind actions. Only DeepSeek-R1 provides access to the reasoning tokens; o3-mini does not provide them, and DeepSeek-V3 is not a reasoning model. 
We conducted an automated log analysis and performed a deep dive into some interesting cases. 

Every model starts with a synthesis action (100\% for DeepSeek-V3, $\sim$90\% for DeepSeek-R1 and o3-mini). The following excerpt illustrates this behavior:
\begin{excerpt}{DeepSeek-R1}
Given the confusion, perhaps the best first step is to synthesize a configuration to see what the actual area and latency are. For example, select the lowest latency options for each function and see what the total area is. So the first request would be to synthesize: fill\_matrix option 2 (lowest latency) reverse\_string option 1 (lowest area among low latency) traceback option 0 (lowest latency) needwun option 0 (only option). 
\end{excerpt}

From DeepSeek-R1's reasoning tokens, the model tends to overanalyze and become confused, often prompting a synthesis to evaluate the outcome.

Next we explore how DeepSeek-R1 formulates ILP problems for 
full-system optimization.
As noted before, DeepSeek-R1 never inspects the code (see also \autoref{fig:actions}), limiting its ability to identify data dependencies, infer parallelism and model latency. 
\begin{excerptWithLatency}{DeepSeek-R1}
Assuming that the total area is the sum of all selected function options' areas, and the total latency is the sum of their latencies (though in reality, it's more complex due to parallelism and pipelining).
\tcblower
\textbf{Latency:} $\text{C6} + OddFact + Fib + 2 * EvenSum$ \\
\textbf{Correct:} $\text{C6} + \max(OddFact, Fib) + 2*EvenSum$
\end{excerptWithLatency}

Indeed, as seen above, DeepSeek-R1 acknowledges that the problem may be more complex but ultimately chooses not to account for parallelism. However, it does recognize the two calls to EvenSum. 
\begin{excerptWithLatency}{o3-mini}
We will compute a "design metric" for each combination as follows:\\
overall\_lat = c6\_laty + delta\_lat\_child\\
overall\_area = c6\_area + delta\_area\_child\\
For delta, we consider the overhead due to calls:
compute6 calls: odd\_factorial (once), fibonacci (once), and even\_sum twice.
We assume the effective latency overhead is the maximum latency among these calls.
\tcblower
\textbf{Latency:} C6 + max(OddFact, Fib, EvenSum) \\
\textbf{Correct:} C6+ max(OddFact, Fib) + 2*EvenSum
\end{excerptWithLatency}
In SYN6, o3-mini has the right idea but fails to accurately model the total latency. Although incorrect, these formulations guide the optimizer by minimizing the latency of the child functions. The LLMs occasionally inferred correct information. For example, DeepSeek-R1 correctly infers that functions called multiple times must use the same configuration:

\begin{excerpt}{DeepSeek-R1}
The call graph shows compute4 calls two even\_sum instances, but the options are per function, not per instance. So even if there are two instances of even\_sum in compute4, they both use the same function's option. Selecting even\_sum's option once applies to all instances.
\end{excerpt}

Other times, the reasoning does not follow design sense:

\begin{excerpt}{DeepSeek-V3}
Let's revisit the configuration with `iir 1` and `fir 0`, which achieved a latency of 8 and an area of 4784.2. Although the area is below the target, the latency is significantly better. To balance this, let's try increasing the area slightly by using `iir 1` and `fir 1`, which previously achieved a latency of 11 and an area of 6019.0, closer to the target.
\end{excerpt}

The model favors worse latency and area because the area is too low compared to the target. Any engineer would understand that a higher area should only be favored if it is below the target and paired with better latency.

We also observe a trend in DeepSeek-V3's action calls. As shown in \autoref{fig:actions}, it performs many more synthesis and inspect calls compared to DeepSeek-R1 and o3-mini. Log analysis reveals a pattern: DeepSeek-V3 synthesizes a design, inspects alternative functions, and, if promising, resynthesizes with the updated functions. This trial-and-error approach continues until all viable combinations are explored. DeepSeek-V3 only chooses to formulate an ILP problem if the trial-and-error approach fails to meet the target.


\section{Related works}
C2HLSC~\cite{c2hlsc, C2HLSC2} is an automatic repair and optimization framework for HLS based on an LLM feedback loop. We employ a similar feedback loop for pragma insertion in task~\ding{202}, but we iterate multiple times, providing synthesis feedback, whereas C2HLSC iterates only once. The LLM then performs task~\ding{203}, finding kernels that satisfy the constraints.

\setlength{\tabcolsep}{2pt}
\begin{table}[htb]
\caption{Area ($\mu m$) and latency (cycles) comparison with manual and previous work results on overlapping benchmarks.}
\label{tab:comp_c2hlsc}
\resizebox{\columnwidth}{!}{
\begin{tabular}{@{}lrrrr|rrrrrr|rr@{}}
\toprule
 & \multicolumn{4}{c|}{C2HLSC~\cite{C2HLSC2}} & \multicolumn{6}{c}{This work} & \multicolumn{2}{|c}{Manual}\\ \cmidrule(l){2-11} 
 & \multicolumn{2}{c}{Sonnet3.5} & \multicolumn{2}{c|}{GPT4-o} & \multicolumn{2}{c}{DeepSeek-V3} & \multicolumn{2}{c}{DeepSeek-R1} & \multicolumn{2}{c}{o3-mini} & \multicolumn{2}{|c}{Impl.}\\ \cmidrule(l){2-13} 
Benchmark & Area & Lat. & Area & Lat. & Area & Lat. & Area & Lat. & Area & Lat. & Area & Lat.\\ \midrule
AES & 2965 & 853 & 2975 & 853 & \textbf{2622} & \textbf{239} & 2665 & 1029 & 2630 & 1096 & 3386& 193 \\
SHA256 & 41924 & 83 & 41794 & 83 & \textbf{37894} & \textbf{67} & 41743 & 84 & \textbf{37894} & \textbf{67}& 36090 & 48 \\
Present &  22245 & \textbf{897} & 19985 & 6193 &  14918  & 4347 & \textbf{12799} & 2226 & 13235 & 3697& 12056& 37 \\\bottomrule
\end{tabular}}
\end{table}
\setlength{\tabcolsep}{6pt}

\autoref{tab:comp_c2hlsc} compares the results on benchmarks common with C2HLSC~\cite{C2HLSC2} and provides reference results obtained with a manual implementation. From the results, our agentic approach achieves a better area on these benchmarks, with latency reduced only in the present benchmark. The manual implementations are derived from the same C code, without added pragmas, used in this work. While there is some margin for improvement, it is minimal.
HLSPilot~\cite{hlspilot} optimizes HLS code using LLMs by combining, profiling, and optimizing C code for HLS. This framework is not open, and the paper does not specify the automation level of the flow. The results are presented as runtime achieved on an FPGA. 


\section{Conclusion}

We propose and evaluate an agentic HLS framework using different LLMs, both with and without a built-in reasoning mechanism, to investigate if and how new-generation reasoning models can aid design automation. 
While our framework improves state-of-the-art LLM-based HLS optimization, reasoning models do not outperform non-reasoning models in this task as they do in math and coding benchmarks~\cite{deepseekr1}. This raises a question for future work: \textit{Do these models perform better in these benchmarks due to training contamination? Can we improve the efficacy of reasoning models with ad-hoc prompt engineering?} We \href{https://anonymous.4open.science/r/anonymous-sub-23F5/README.md}{open-source our framework}. 

\newpage
\bibliographystyle{IEEEtran}
\bibliography{main.bib}

\end{document}